\documentclass{article}

\usepackage{xcolor}
\usepackage{graphicx}
\usepackage{float}
\usepackage{subcaption}

\usepackage[final]{
tackling_climate_workshop_style}

\usepackage[utf8]{inputenc} 
\usepackage[T1]{fontenc}    
\usepackage{hyperref}       
\usepackage{url}            
\usepackage{booktabs}       
\usepackage{amsfonts, amsmath}       
\usepackage{nicefrac}       
\usepackage{microtype}      
\usepackage{wrapfig}
\usepackage{todonotes}

\title{Can Reinforcement Learning support policy makers? A preliminary study with Integrated Assessment Models}

\author{Theodore Wolf \\
  University College London | Carbon Re\\ 
  \texttt{theo@carbonre.com} \\
  \And
    Nantas Nardelli \\
   Carbon Re\\
   \texttt{nanta@carbonre.com} \\
   \And
      John Shawe-Taylor \\
      University College London \\
        \texttt{j.shawe-taylor@ucl.ac.uk} \\
      \And
   Mar\'{i}a P\'{e}rez Ortiz \\
   University College London \\
   \texttt{maria.perez@ucl.ac.uk} \\
}

\begin{document}

\maketitle

\begin{abstract}

Governments around the world aspire to ground decision-making on evidence. Many of
the foundations of policy making — e.g. sensing patterns that relate to societal needs, developing
evidence-based programs, forecasting potential outcomes of policy changes, and monitoring effectiveness of policy programs — have the potential to benefit from the use of large-scale
datasets or simulations together with intelligent algorithms. These could, if designed and deployed in a way that is
well grounded on scientific evidence, enable a more comprehensive, faster, and rigorous approach to policy making. 
Integrated Assessment Models (IAM) is a broad umbrella covering scientific models that attempt to link main features of society and economy with the biosphere into one modelling framework. 
At present, these systems are probed by policy makers and advisory groups in a hypothesis-driven manner. 
In this paper, we empirically demonstrate that modern Reinforcement Learning can be used to probe IAMs and explore the space of solutions in a more principled manner.
While the implication of our results are modest since the environment is simplistic, we believe that this is a stepping stone towards more ambitious use cases, which could allow for effective exploration of policies and understanding of their consequences and limitations. 

\end{abstract}

\section{Introduction}
%
%
Climate is a high dimensional dynamical system with strong inter-dependent components and long time dependencies, all of which interact to produce highly non-linear responses and behavior.
Climate is also highly conditioned on human behavior -- another greatly complex system -- such that is now necessary to reason about climate change from a socio-climatic perspective~\citep{Moore2022DeterminantsSystem}.
%
%
To make progress towards achieving some kind of solution in the face of extreme consequences, policymakers and advisory groups employ \emph{Integrated Assessment Models} (IAMs), state of the art models for climate change that combine knowledge about human development (such as economical theories) together with planetary sciences such as ecology and geophysics~\citep{parson1997integrated}.
%
Exploring and analysing the properties of the IAMs employed for large scale assessments~\citep{portner2022climate} - to e.g. measure fidelity against the real world -- is generally intractable from a computational perspective, which leads researchers to implement poor simplifying assumptions and decrease their effectiveness~\citep{asefi2021failure}.
Smaller IAM models aim to provide an alternative by employing fewer state variables and simpler sets of dynamics, making them amenable to mathematical probing and analysis~\citep{Kittel2017FromManagement, Nitzbon2017SustainabilityModel}.
%
The literature commonly explores these models with ODE solvers; however recently \citep{Strnad2019DeepStrategies} has shown that it is possible to employ them as environments in standard Reinforcement Learning (RL) ~\citep{Sutton2020ReinforcementIntroduction}, and explore models using trained policies.
These can be used to understand the system, and provide upwards insight towards improving more complex IAMs or even our understanding of climate change policies.
%

We build on this work, testing more RL algorithms, reward functions, as well as different experimental setups.
Among others, we show that (a) modern RL can learn effective policies with a variety of reward functions in this environment (b) that different agents and reward functions generate a significantly diverse set of solutions, thus exploring the IAM in different manners, and (c) that it is necessary to apply care when designing reward functions as they show different success rate in reaching the desirable state for different initialisation points, finally -  (d)  that RL helps us gain a deeper understanding of the properties and limitations of the applied models.
\section{AYS environment, RL reward functions and agents}
\begin{wrapfigure}{r}{0.45\textwidth}
  \vspace{-5mm}
  \centering
  \includegraphics[width=0.43\textwidth]{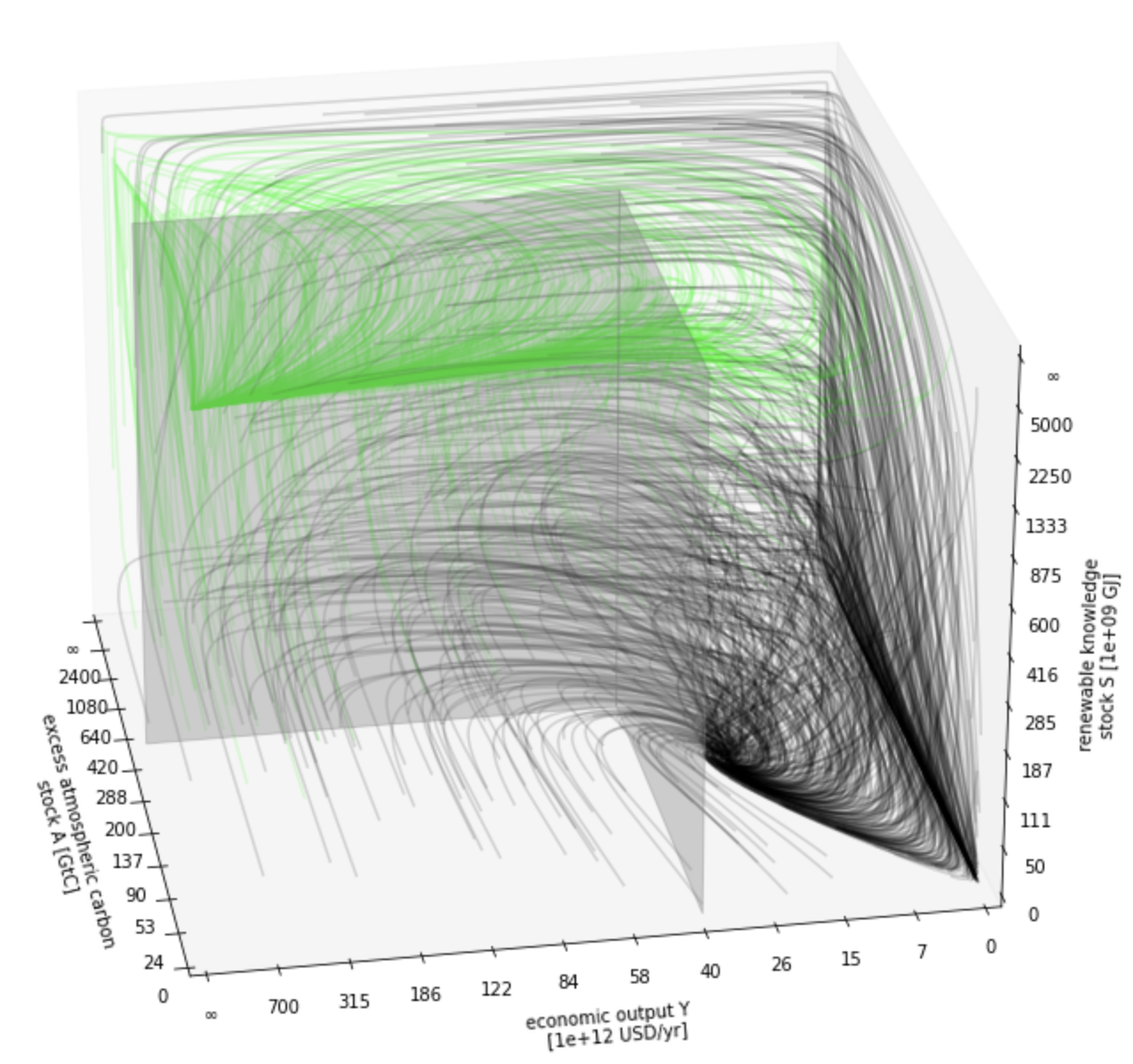}
          \vspace{-4mm}
  \caption{Phase space of the environment, each dimension corresponds to one of the states dimensions. Hair lines show the flows of the system.}
  \label{fig:phase_space}
  \vspace{-5mm}
\end{wrapfigure}

%
%
We employ the \emph{AYS model}~\citep{Kittel2017FromManagement}, which we use to create a Markov Decision Process following \cite{Strnad2019DeepStrategies}.
AYS is governed by three coupled differential equations: 
\begin{equation}
\frac{dA}{dt} = \frac{1}{1+(S/\sigma)^\rho}\frac{Y}{\phi\epsilon} - A/\tau_A, \quad \frac{dY}{dt} = \beta Y - \theta AY,\quad  \frac{dS}{dt} = \left(1-\frac{1}{1+(S/\sigma)^\rho}\right)\frac{Y}{\epsilon} - S/\tau_S,
\end{equation}
where $A$ is the excess atmospheric carbon, $Y$ society's economic output, and $S$ the renewable knowledge stock.
At each step $t$, the agent observes a vector $(A_t, Y_t, S_t)$, which is a partial representation of the state of the system.
This dynamical system contains two attractors, a \emph{green fixed point}, where renewable knowledge and economic output grow forever, and a \emph{black fixed point}, where the economic output is stagnant and there is a large amount of excess carbon in the atmosphere (Figure~\ref{fig:phase_space}).
Broadly speaking, the former is a positive end state (and effectively encodes the high level goal), while the latter is undesirable.
See Appendix~\ref{appendix:AYS} for a more detailed technical aspects of the environment.
The agent can take four actions, each corresponding to a high level government policy decision and changes to the dynamics of the environment (See Table~\ref{tab:actions} for a mathematical and textual description of the actions).
\paragraph{Reward functions}
The reward signal that agents receives from taking actions in this environment is based on \emph{planetary boundaries} (PBs), quantitative physical and economical limits which, if crossed, would represent disastrous and irreversible consequences for the biosphere and humans \citep{Rockstrom2009AHumanity, Steffen2015PlanetaryPlanet}.
We define these boundaries as $A_{PB}=600$, $Y_{PB}=4\times10^{13}$, and $S_{PB}=0$, which form the triplet $s_{PB}$.
Crossing one of these boundaries yields a reward of 0 and ends the episode.
Furthermore, in the standard scenario, the agent is rewarded for staying away from these boundaries, i.e. $R_t^{PB} =  ||s_t-s_{PB}||^2$. This incentives the agent to maximise economic output while minimising carbon emissions.  We call this reward function \emph{PB reward}. 
A second reward function we employ is the  \emph{Policy Cost (PC) reward}, which adds an action-dependent cost to PB reward, simulating the real world cost of implementing and maintaining any significant shift in policy in a running socio-economical system. E.g., throttling growth over years may be challenging for a policymaker~\citep{keysser20211}.
Finally, our experiments see the use of a third and simpler reward function (\emph{Sparse reward}, since it significantly lowers the amount of feedback
\newpage that the agent receives on average), which only considers whether the agent reaches the goal or hits any of the planetary boundaries:

\vspace{-4mm}
\small
\begin{align}
    r(s) = 
    \begin{cases}
    1 \quad \textrm{if} \quad s_t=s_g\\
    -1 \quad \textrm{if} \quad (A_t>A_{PB}) \vee (Y_t<Y_{PB})\\
    0 \quad \textrm{otherwise}.
    \end{cases}
\end{align}

\normalsize
\vspace{-3mm}

\begin{table}[ht]
\vspace{-5mm}
\centering
\caption{Environment actions, and how they relate to the policy cost (PC) reward function.}
\scriptsize
\begin{tabular}{ccll}
\toprule
\textbf{Action} & Parameter Change & Explanation & PC reward\\ \midrule
Noop  & None                                                     & Environment evolves with default parameters. & $R_t^{PB}$\\
ET    & $\sigma \gets \sigma/\sqrt{\rho}$                        & Halves relative cost of renewables to fossil fuels. & $0.5\times R_t^{PB}$\\ 
DG    & $\beta \gets \beta/2$                                    & Halves rate of growth of the economy. & $0.5\times R_t^{PB}$\\
ET+DG & $\beta \gets \beta/2$, $\sigma \gets \sigma/\sqrt{\rho}$ & Combination of the two above actions. & $0.25\times R_t^{PB}$\\
\bottomrule
\end{tabular}
\label{tab:actions}
\end{table}

\vspace{-3mm}
\paragraph{Agent settings} 
We are primarily interested in the interplay of RL and AYS. To do so, we learn a diverse set of agents through four different learning algorithms: A2C~\citep{Mnih2016AsynchronousLearning}, PPO~\citep{Schulman2017ProximalAlgorithms}, DQN~\citep{Mnih2015Human-levelLearning}, and Double DQN with dueling networks and prioritized experience replay (D3QN)~\citep{Hessel2017Rainbow:Learning}.
All agents employ the same network architecture, bar the final layer, which is dependent on the number and type of heads required by the algorithm. See Appendix~\ref{appendix:network} for further details on the network architecture.
Agents are trained for 5e5 steps, and tuned separately for best cumulative performance using Bayesian optimization. All experiments show results of 3 random seeds.
\section{Results and discussion}
\begin{figure}[h]
\vspace{-3mm}
    \centering
    \includegraphics[width=\textwidth]{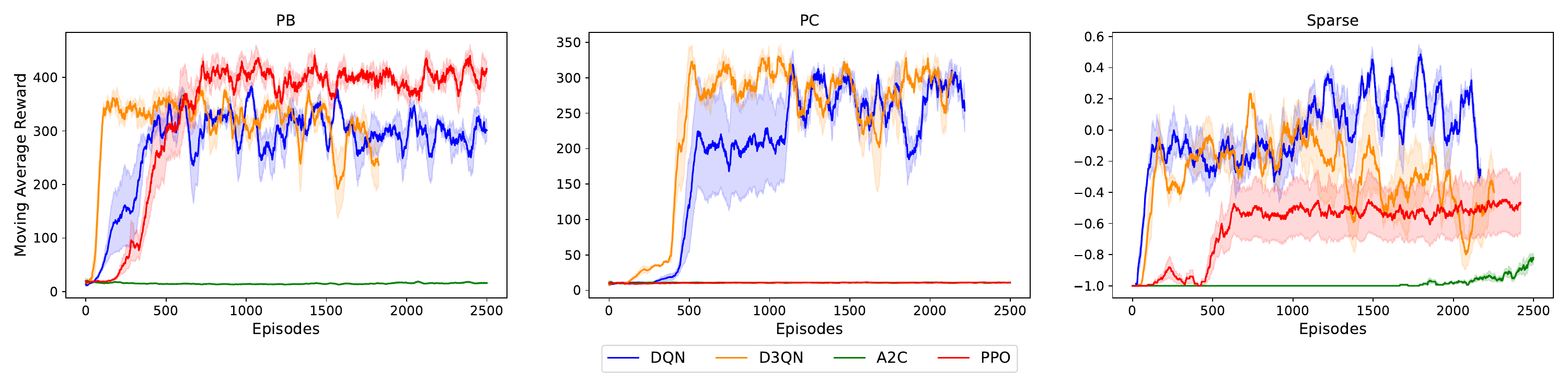}
            \vspace{-5mm}
    \caption{Average cumulative training rewards for each agent by reward function.}
    \label{fig:training}
\end{figure}

Figure~\ref{fig:training} shows how most agents are generally able to learn a policy that optimizes the relevant reward function. Interestingly, while DQN and D3QN are broadly consistent across reward functions, A2C struggles to learn policies under the same episodic budget, even under a substantial amount of hyperparameter tuning.
PPO on the other hand beats all agents on \emph{PB reward}, and underperforms everywhere else.

\vspace{-3mm}

We hypothesize these differences may be related to the use of experience replay for the DQN agents, as well as the action sampling method in the case of A2C and PPO.
We will later see that a robust class of successful policies in this environment corresponds to getting as quickly as possible to a spot under a green "current", and letting the system converge to the goal.
In such scenario, the action distribution for large subsets of the state space needs to converge onto specific actions (such as \emph{noop} in this particular case).
This is a straightforward outcome for DQN-based agents, however as PPO relies on softmax parametrization to output actions (and learn) in a discrete space, it becomes more difficult for the agent to converge to such a policy given its dependency on both entropy regularization and Boltzmann-based exploration~\citep{ahmed2019understanding, mei2020global}.
%

%
\begin{figure}[h]
\vspace{-3mm}
    \centering
    \begin{subfigure}[b]{0.325\textwidth}
        \includegraphics[width=\textwidth]{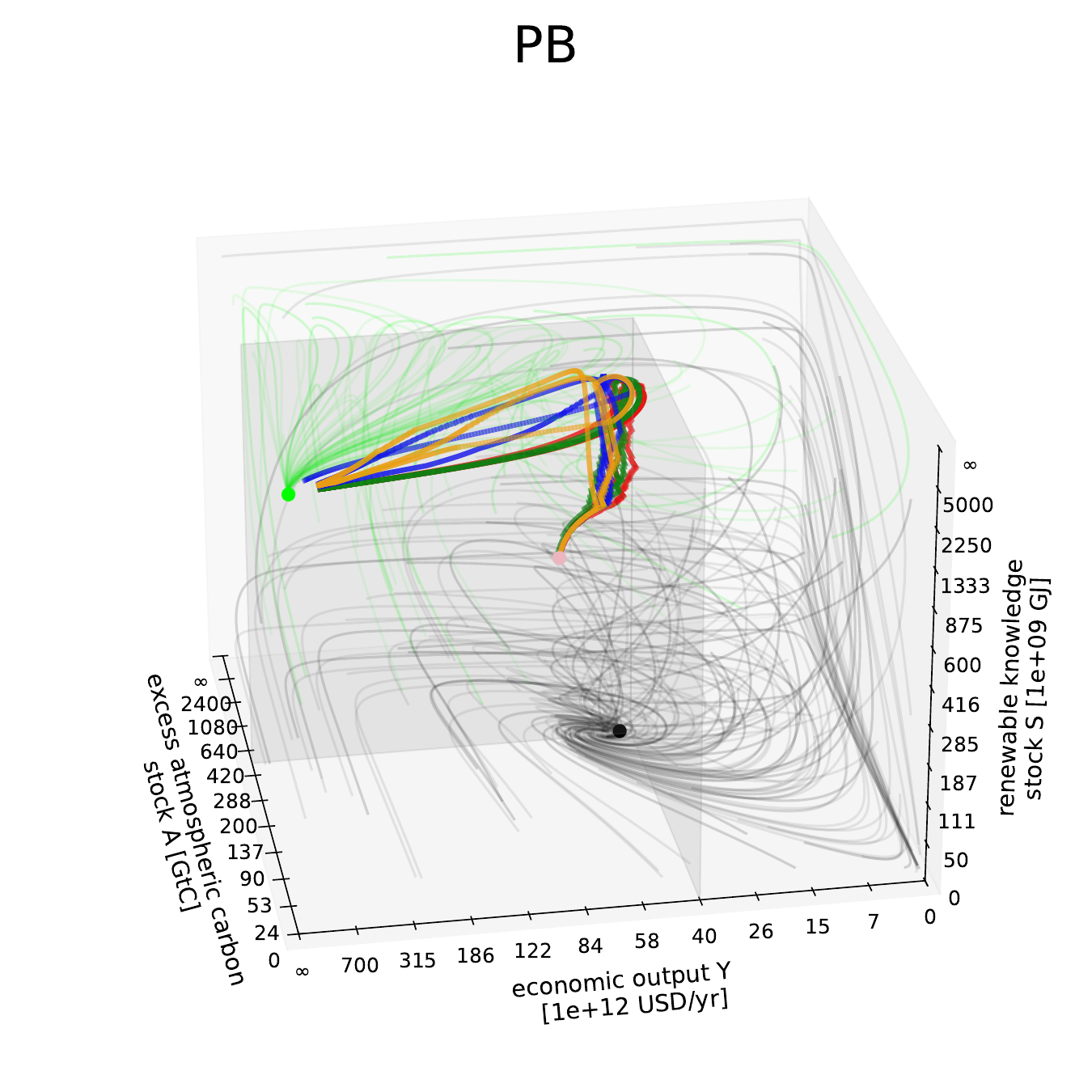}
        \label{fig:d3qntraj}
    \end{subfigure}
    \begin{subfigure}[b]{0.325\textwidth}
        \includegraphics[width=\textwidth]{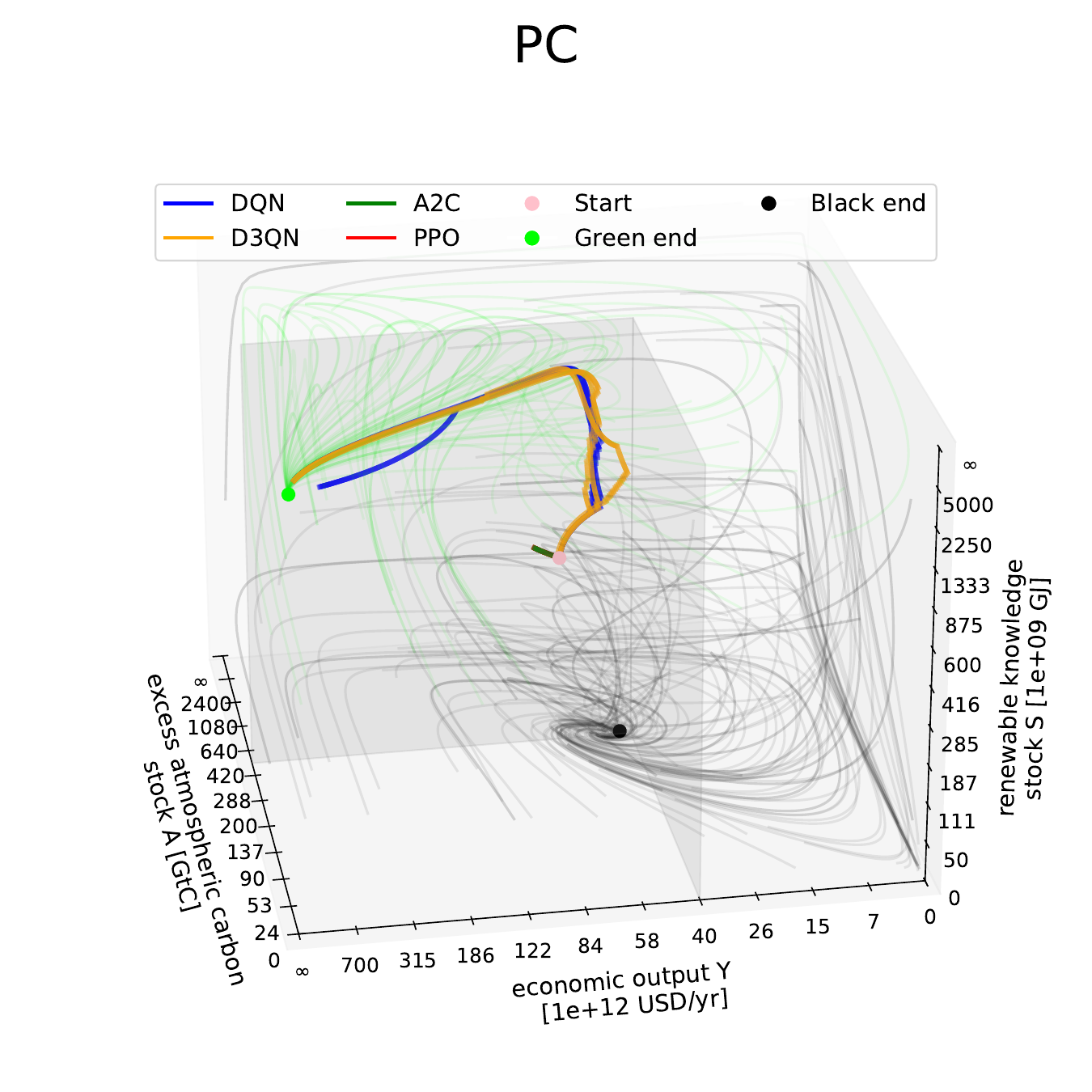}
        \label{fig:ddqntraj}
    \end{subfigure}
    \begin{subfigure}[b]{0.325\textwidth}
        \includegraphics[width=\textwidth]{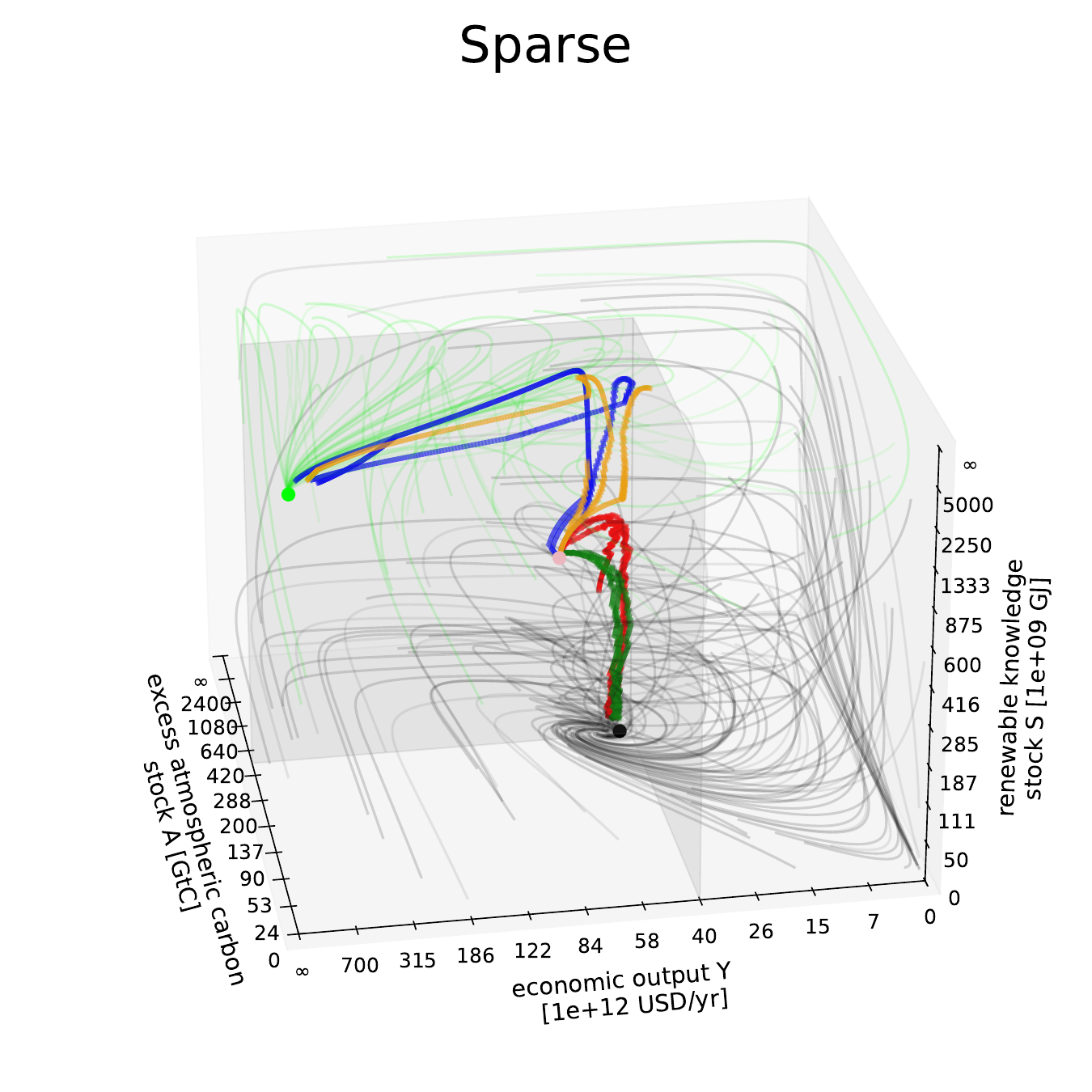}
        \label{fig:ppotraj}
    \end{subfigure}
            \vspace{-5mm}
    \caption{Sample trajectories of the four agents initialized from a fixed state with 3 reward functions. The PB reward function helps the agents solve the environment more consistently.}
    \label{fig:trajs}
\end{figure}
Figure~\ref{fig:trajs} shows that agents converge to significantly different policies under different learning algorithms and reward functions. 
We note that for a significant part of the initial state space, the first few actions for all the agents that reach the green point are generally similar. 
\begin{figure}
\vspace{-3mm}
    \centering
    \begin{subfigure}[b]{0.325\textwidth}
        \includegraphics[width=\textwidth]{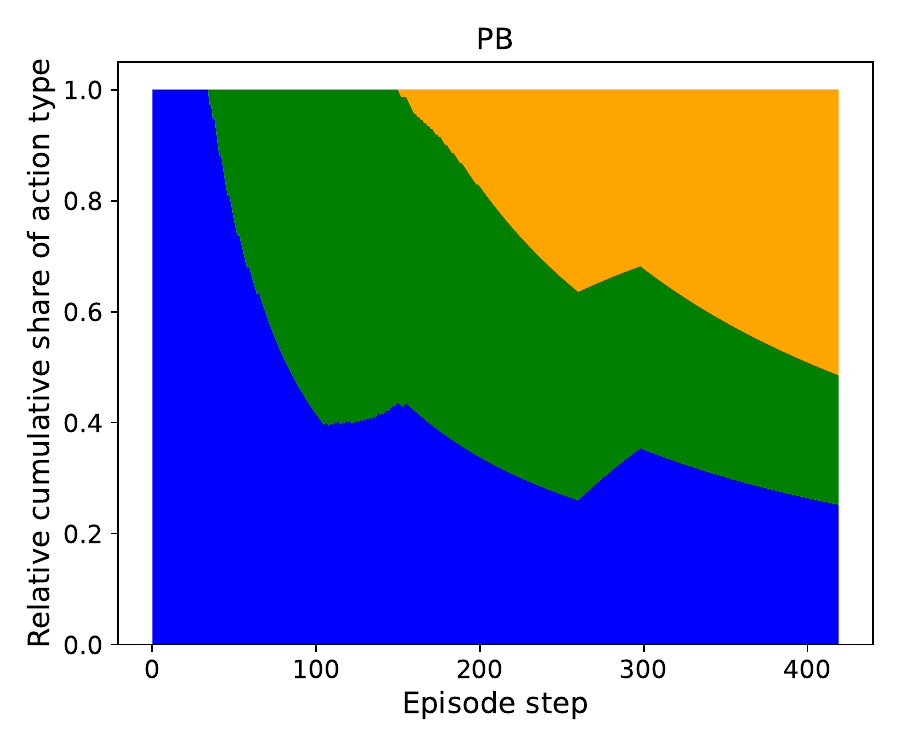}
        \label{fig:sparse_pb}
    \end{subfigure}
    \begin{subfigure}[b]{0.325\textwidth}
        \includegraphics[width=\textwidth]{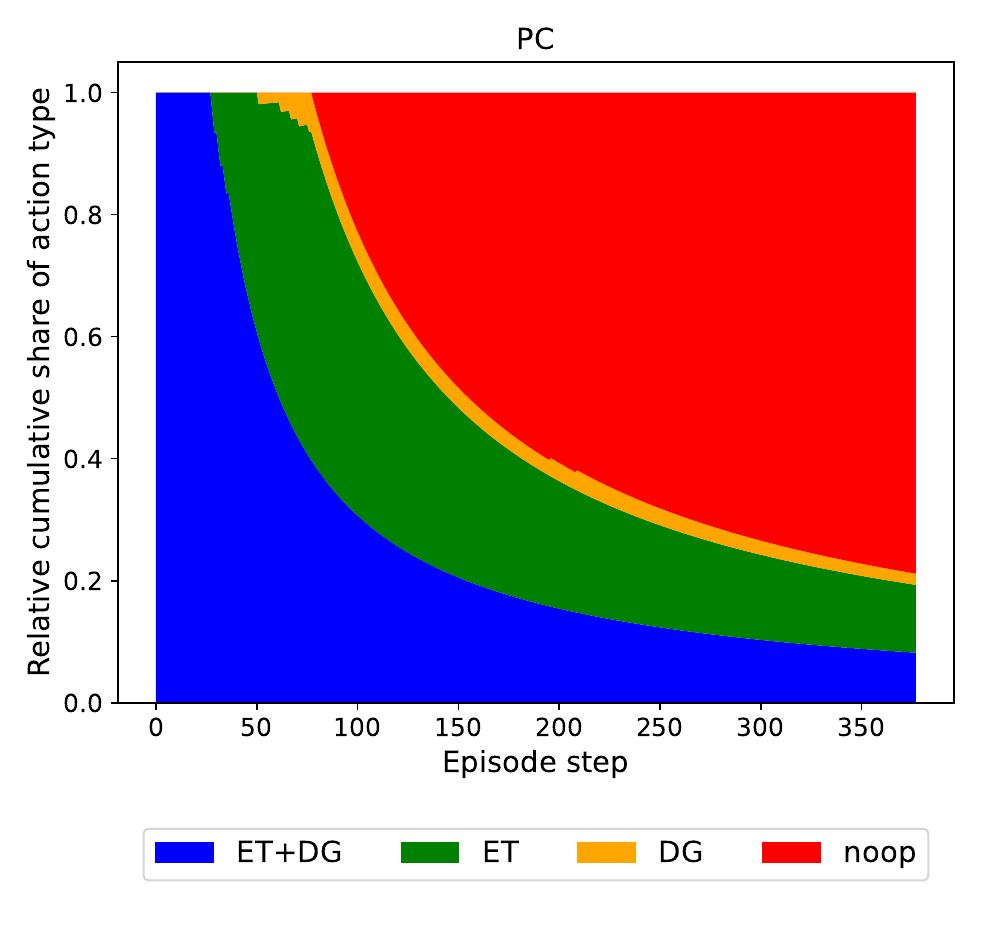}
        \label{fig:sparse_pc}
        \vspace{-5mm}
    \end{subfigure}
    \begin{subfigure}[b]{0.325\textwidth}
        \includegraphics[width=\textwidth]{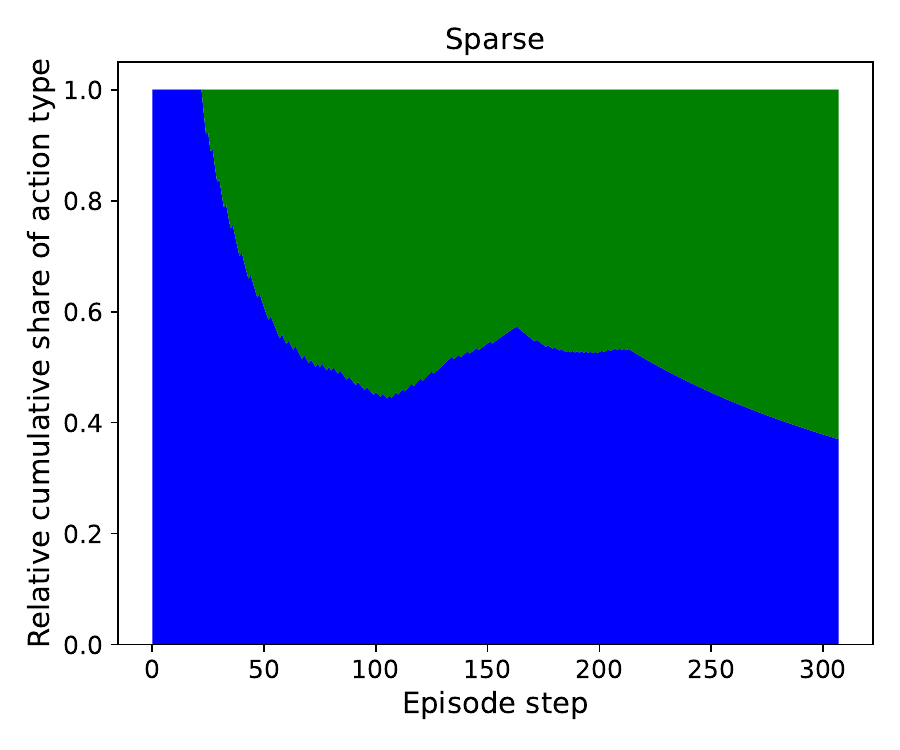}
        \label{fig:sparse_eq}
    \end{subfigure}
            \vspace{-3mm}
    \caption{Sample cumulative relative share of actions taken by DQN agents trained with different reward functions.}
    \label{fig:equity}
    \vspace{-5mm}
\end{figure}

In Figure \ref{fig:equity}, we see that agents that reach the green point prefer picking the action \emph{ET+DG} during the first 23 steps of the episode, and then go on to exhibiting more variety later in the episode. We notice the significant difference between the distribution of actions for the different reward functions. As expected under the PC reward, the agent maximises the number of \textit{noop} actions taken in order to collect the highest reward. What is important to note is that the agent prefers taking on the cost of taking actions in the beginning of the episode. More generally, we see that the distribution of the first 50 actions taken are the same across all reward functions (75\% \textit{DG+ET} and 25\% \textit{ET}).

This implies that the environment has some form of natural bottleneck, which can also explain the loss in general performance for some of the agents in the sparse reward case, since agents don't have a state or dynamics-driven exploration system~\citep{burda2018exploration}.
Ultimately, we see that the different reward functions yield qualitative changes in the resulting trained policy of the agents, with a generally higher difference for on-policy agents. 
\begin{figure}[b]
\vspace{-3mm}
    \centering
    \begin{subfigure}[b]{0.325\textwidth}
        \includegraphics[width=\textwidth]{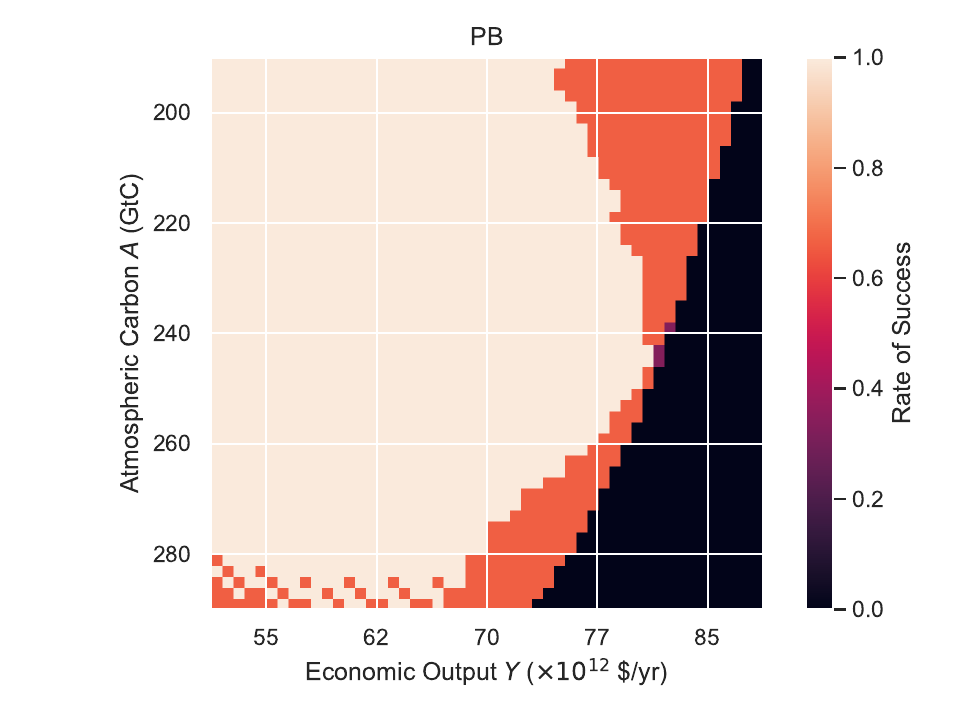}
        \label{fig:pb_end}
    \end{subfigure}
    \begin{subfigure}[b]{0.325\textwidth}
        \includegraphics[width=\textwidth]{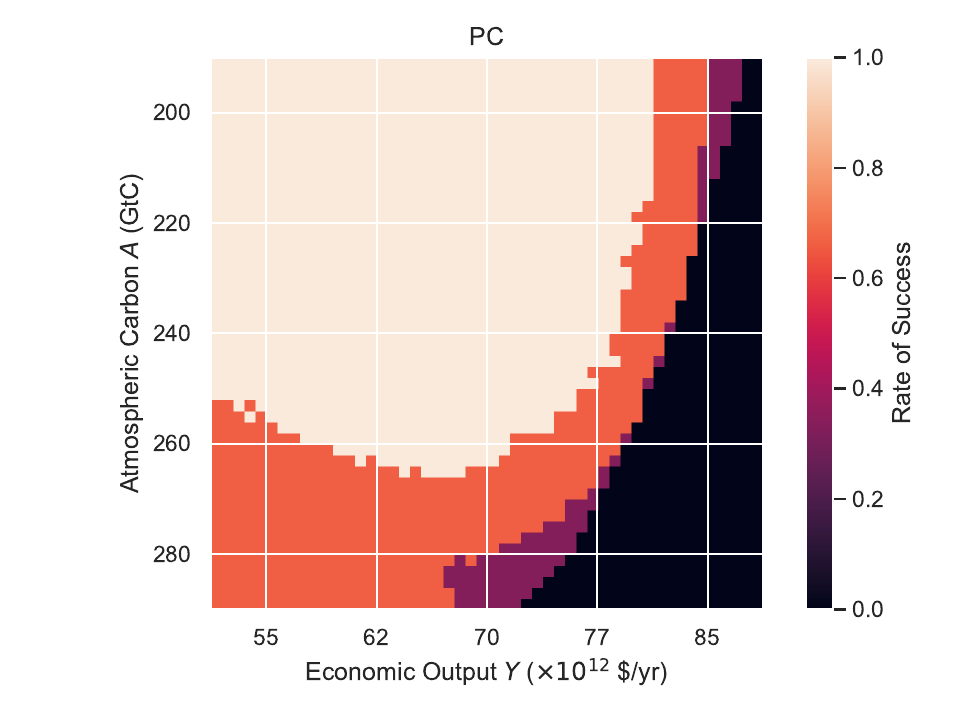}
        \label{fig:pc_end}
    \end{subfigure}
    \begin{subfigure}[b]{0.325\textwidth}
        \includegraphics[width=\textwidth]{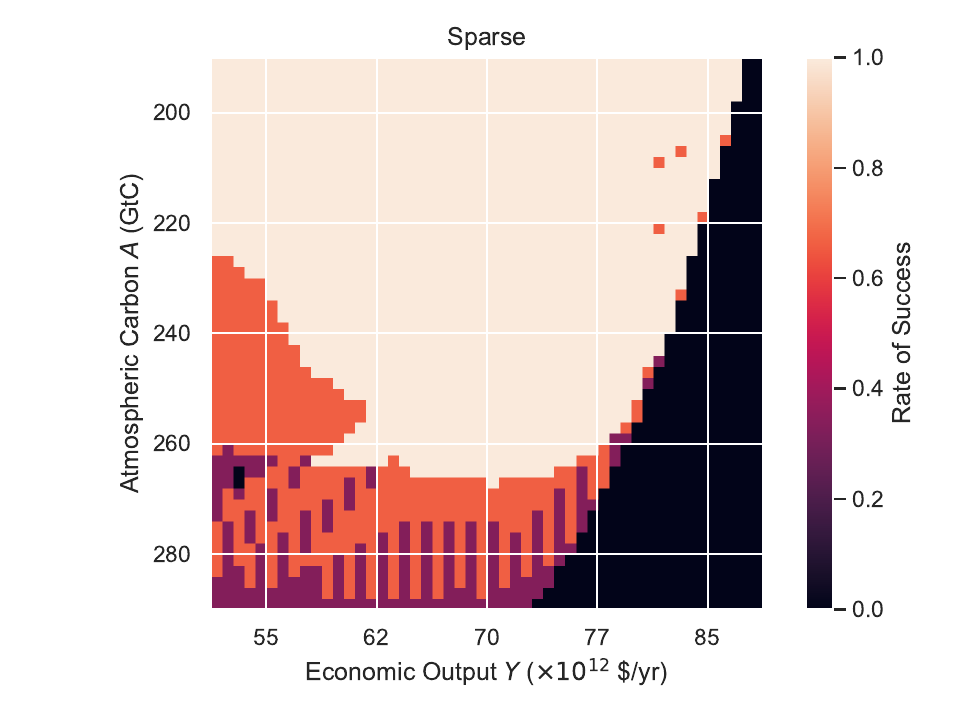}
        \label{fig:sparse_end}
    \end{subfigure}
            \vspace{-5mm}
    \caption{Average success rate of three seeds of the trained DQN agent for different reward functions given different initialisation states. Success is defined as reaching the green point.}
    \label{fig:success}
    \vspace{-3mm}
\end{figure}

Figure~\ref{fig:success} shows how agents solve the environment from different initial states when trained with different reward functions. 
The experiments suggest that policies are indeed highly dependent on the employed learning algorithm, but we also see commonalities across reward functions, particularly when looking at initial states.
If the agents are initialized in the top-left of the starting grid, is is much easier for them to reach the goal.
On the other hand if they start in the bottom-right corner, they seem guaranteed to fail. 
Therefore when initialising a new episode, not all starts are equal in outcome. We hypothesize that a strictly optimal policy may not be able to do better, but further work is necessary to establish the exact failure conditions.
We note that under this model, we can explain this particular result by looking at the modeled dynamics of real the environment: the economic output grows exponentially (as long as atmospheric carbon is low) and needs exponentially more energy to then sustain itself. 
The underlying equations dictate that if not enough renewable knowledge stock is available, then this energy will come from fossil fuels. 
This consequently increases the excess atmospheric carbon which accumulates very fast and causes the agent to cross the atmospheric carbon boundary.
Such emergent behavior is also not immediately obvious when considering the system equations.

%

\paragraph{Conclusions}
The question of whether algorithms could support policy making is of current interest, where more than ever we find governments having to make high-stake decisions with regards to fast changing complex, global and interconnected challenges, which are difficult to understand and tackle without relevant datasets, scientific evidence and scenario-analysis tools.
Public policy-making is often a cyclical process, with stages such as identification of societal needs, formulation of agendas, scenarios and policy alternatives, adoption of policy decisions, implementation in the real world, and finally, evaluation of their effectiveness, with subsequent improvements.  
Our experiments aim to understand whether RL can be used to formulate policy alternatives and evaluate their effectiveness in a simulation environment. 
Specifically, we have shown that a variety of RL algorithms produce well-behaved policies in the AYS environment under different, more or less sparse, reward functions. 
The combination of different reward functions, RL algorithms, and the space of initial states produce diverse policies that can successfully explore and solve the underlying AYS model. 
This enables the study of the emergent properties of the system without needing to encode much knowledge about the model into the learning process. This is an interesting result, as it shows the potential of applying RL as a general debugging and analysis toolkit for IAM models. 

We note that the solution to solving the AYS model relates to \textit{early action} \citep{MacCrakenEarly}, as the agent solves the environment more consistently when implementing the more aggressive policy position early on in the episode. Early action dictates that fast and aggressive climate change mitigation has cumulative benefits. The concept of \emph{early action} stems from the \emph{time value of carbon} \citep{CORNELISVANKOOTEN2021106162}.
%
Future work should look into exploring whether agents can be trained across multiple model-environments, to understand whether some kind of ``common exploration strategy'' emerges as a result, or whether agents could be trained to explore small, simplified models, and behave in a reasonable manner in computationally bigger, and thus more expensive IAMs.
Appendix \ref{appendix:fexperiments} presents further experiments, conclusions and future work.


\bibliography{ref}

\appendix 
\newpage

\section{A broader look at the AYS environment}
\label{appendix:AYS}

\begin{figure}[h]
    \centering
    \includegraphics[scale=0.6]{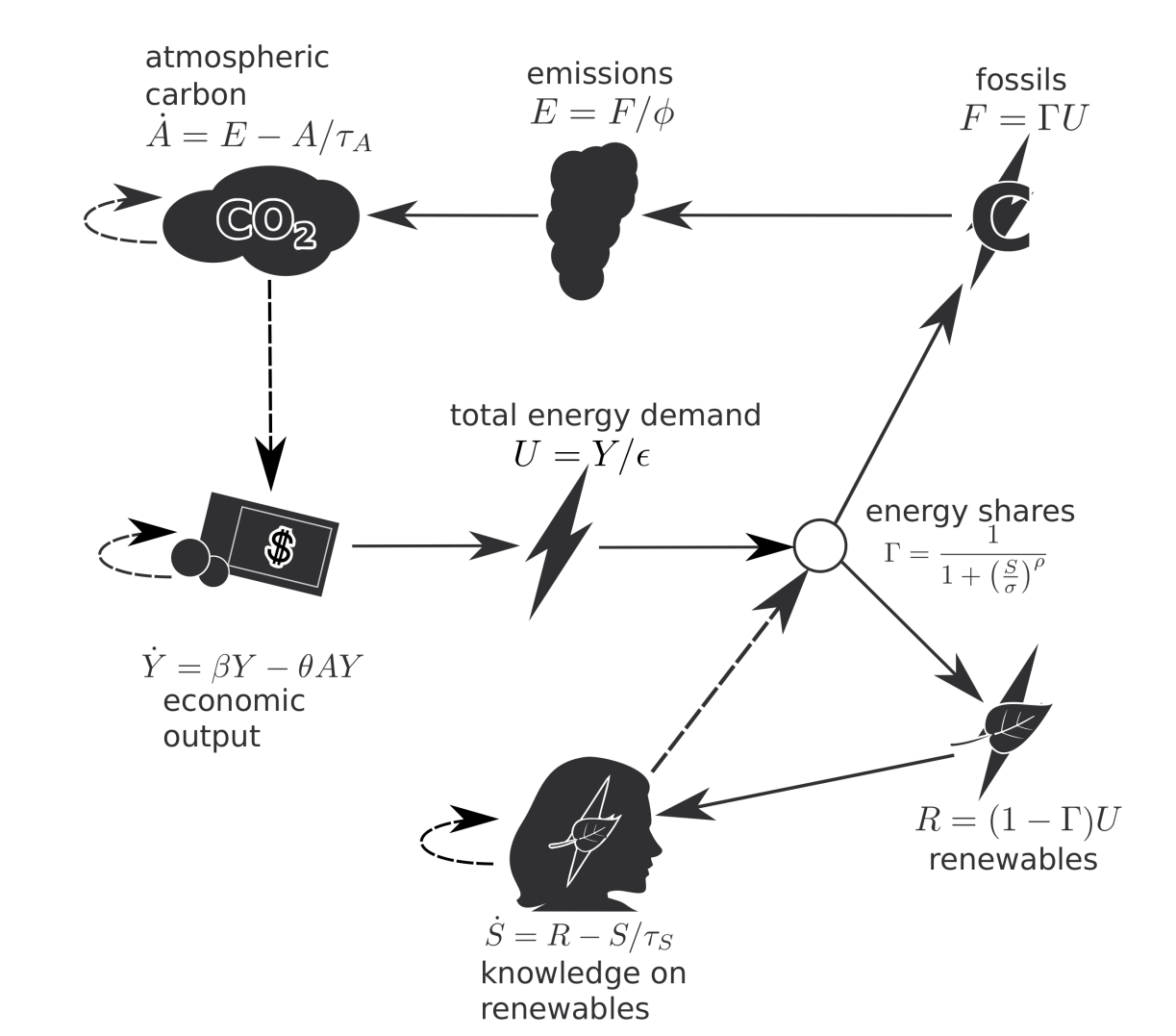}
    \caption{Schematic from \cite{Kittel2017FromManagement} summarizing the interactions in the AYS model. The full lines are positive interactions and the dotted lines are negative interactions.}
    \label{fig:interactions}
\end{figure}

\subsection{Observables}
The model has three observed variables: 
\begin{itemize}
    \item Excess atmospheric carbon, $A$, in Gigaton of Carbon (GtC).
    \item Economic output, $Y$, in US dollars per year (\$/yr).
    \item Renewable knowledge stock, $S$, in GigaJoules (GJ).
\end{itemize}
This low dimensional environment enables us to test our framework's limits as well as plot trajectories in phase space for tractable interpretability and analysis.
Furthermore, the dynamics of model are also kept relatively simple compared to other more complex models \cite{Nitzbon2017SustainabilityModel, Bury2019ChartingModel,Moore2022DeterminantsSystem}.
We note that out of all three variables, the last one is less "tangible" than other ones, as it represents some knowledge metric humans have about renewable energy; to make it relevant to available quantitative data, this model chooses to represent it as energy.

\subsection{Equations}
The system is governed by three differential equations, one for each observed variable:
\begin{align}
    \frac{dA}{dt} &= E- A/\tau_A,\\
    \frac{dY}{dt} &= \beta Y - \theta AY,\\
    \frac{dS}{dt} &= R - S/\tau_S.
\end{align}
With $R$ and $E$ the energy extracted from renewables and fossil fuels respectively. These are defined with the energy demand $U$ in GJ/year, which is proportional to the economic output:
\begin{align}
    U=\frac{Y}{\epsilon},
\end{align}
where $\epsilon$ is the efficiency of energy. Energy is either produced from renewable sources or from fossil fuel sources: 
\begin{align}
    R=(1-\Gamma)U,\\ F=\Gamma U,\\ E = F/\phi.
\end{align}
Here, $\phi$ is the fossil fuel combustion efficiency in GJ/GtC. The share of fossil fuel energy $\Gamma$ is calculated as an inverse response to the renewable knowledge: 
\begin{align}
    \Gamma = \frac{1}{1+(S/\sigma)^\rho}.
\end{align}
With $\sigma$ being the break-even knowledge, which corresponds to the state where renewable and fossil fuel costs become equal, and $\rho$ is the renewable knowledge learning rate. As knowledge on renewables ($S$) increases, $\Gamma \to 0$ and the total energy share produced by renewables increases. If $S\to 0$, then $\Gamma \to 1$ and more energy is produced from fossil fuels.
The interactions are summarized in Figure \ref{fig:interactions}. The parameter values are summarized in Table~\ref{table:params}.

\begin{table}[H]
\centering
\caption{Table summarising the parameters of the AYS model from \cite{Kittel2017FromManagement}.}
\begin{tabular}{cll}
\toprule
\textbf{Parameter} & \textbf{Value} & \textbf{Description} \\ \midrule
\textbf{$\tau_A$} & 50 years & Carbon decay out of the atmosphere. \\
\textbf{$\tau_S$} & 50 years & Decay of renewable knowledge stock. \\ 
\textbf{$\beta$} & 3 \%/year & Economic output growth. \\ 
\textbf{$\sigma$} & $4\times 10^{12}$ GJ & \begin{tabular}[c]{@{}l@{}}Break-even knowledge: knowledge at which \\ fossil fuel and renewables have equal cost.\end{tabular} \\ 
\textbf{$\phi$} & $4.7\times 10^{10}$ GJ/GtC & Fossil fuel combustion efficiency. \\ 
\textbf{$\epsilon$} & 147 \$/GJ & Energy efficiency parameter. \\ 
\textbf{$\theta$} & $8.57\times10^{-5}$ & Temperature sensitivity parameter. \\
\textbf{$\rho$} & 2 & Learning rate of renewable knowledge. \\ 
\bottomrule
\end{tabular}
\label{table:params}
\end{table}

\subsection{States}
The initial state is:
\begin{align}
s_{t=0} = 
    \begin{pmatrix}
    240\, GtC\\
    7\times 10^{13}\,\$/yr\\
    5\times10^{11}\, GJ\\
    \end{pmatrix},
\end{align}
which aims to represent the current state of the Earth in this model.
There are two attractors in this model, 
\begin{align}
s_b = 
    \begin{pmatrix}
    \beta/\theta\\
    \frac{\phi\beta\epsilon}{\theta \tau_A}\\
    0\\
    \end{pmatrix}
    =
    \begin{pmatrix}
    350\, GtC\\
    4.84\times10^{13}\,\$/yr\\
    0\, GJ
    \end{pmatrix}.
\end{align}
This is denoted as the \textit{black fixed point}: roughly half of the current economic production, and this economic production is stagnant.
Furthermore, in the black fixed point, there is an excess of 350 GtC in the atmosphere with no renewable energy production.
The other point we are interested in is located at the boundaries of the state space,
\begin{align}
    s_g = 
    \begin{pmatrix}
    0\\
    +\infty\\
    +\infty
    \end{pmatrix},
\end{align}
where economic growth and renewable energy knowledge grow forever. We label this the \textit{green fixed point}, and we note that it corresponds to the ideal scenario.
The dynamics of this environment do not allow for more fixed points. Therefore, any point in the space will be naturally drawn to one of these fixed points.

Just as in \cite{Kittel2017FromManagement}, we normalize the state variables $A$, $Y$ and $S$ between 0 and 1. 
This prevents any numerical issues from arising in unexpected ways.
The normalization scheme employed is the following:
\begin{align}
    \Bar{s_t} = \frac{s_{t}}{s_t + s_{t=0}}.
\end{align}
This leads to the initial state being $(0.5, 0.5, 0.5)^T$ and the green fixed point to be $s_g=(0, 1, 1)^T$.

\subsection{Episode Description}
The AYS model is a deterministic environment. The enable non-trivial learning, we initialize each new episode n a random state sampled under a fixed distribution:
\begin{align}
s_{t=0}=\begin{pmatrix}
0.5 + \mathcal{U}(-0.05, 0.05)\\
0.5 + \mathcal{U}(-0.05, 0.05)\\
0.5
\end{pmatrix},
\end{align}
where $\mathcal{U}$ is the uniform distribution.
We do not add noise the third varialbe, as we notice that it dramatically reduces the ability for the agent to learn \cite{Strnad2019DeepStrategies}.
%

Each step corresponds to a difference of 1 year. The environment uses an Ordinary Differential Equation numerical solver to calculate the state for the next step given the action-specific parameters.

\section{Neural Network Architecture}
\label{appendix:network}

We attempt to normalize the network architecture across all our experiments (and agents). 
The "torso" of the policy corresponds to three linear layers intercalated with ReLU activation functions.
The input layer has three units, the hidden layer has 256 units and the output layer has four units. 
In the case of D3QN, we use a dueling network~\citep{Wang2015DuelingLearning} such that there are two output layers connected to the hidden layer: one with four units and one with a single unit.

\section{Action distribution by reward function}\label{appendix:plots}

We observe how different reward functions produce significantly different types of trajectories (Figure~\ref{fig:success}).
\begin{figure}[H]
    \centering
    \begin{subfigure}[b]{0.32\textwidth}
        \includegraphics[width=\textwidth]{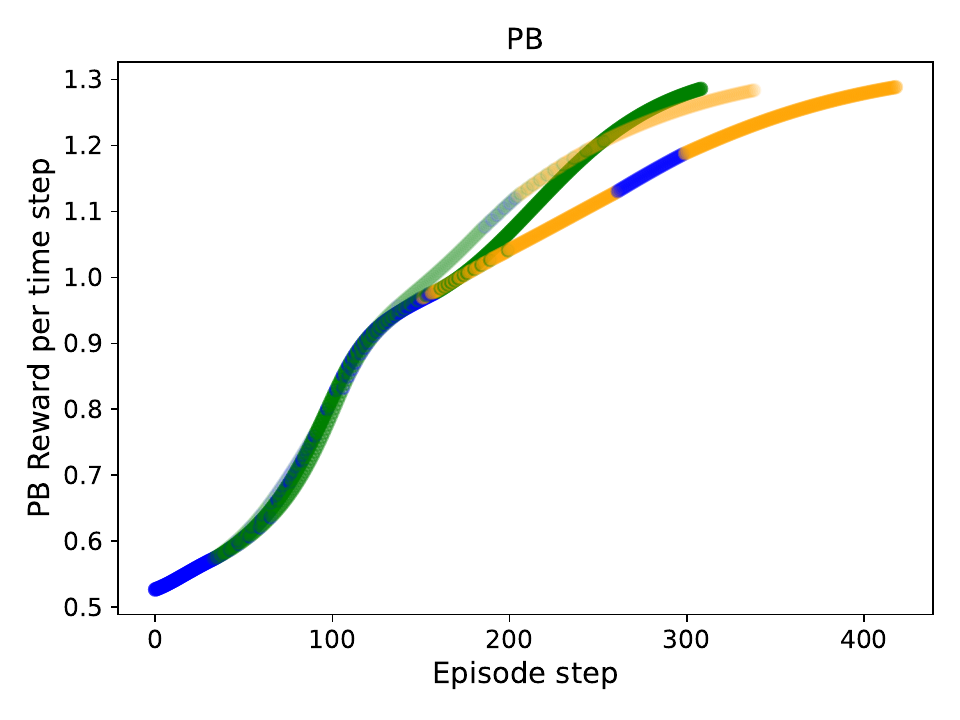}
        \label{fig:pb_ra}
    \end{subfigure}
    \begin{subfigure}[b]{0.32\textwidth}
        \includegraphics[width=\textwidth]{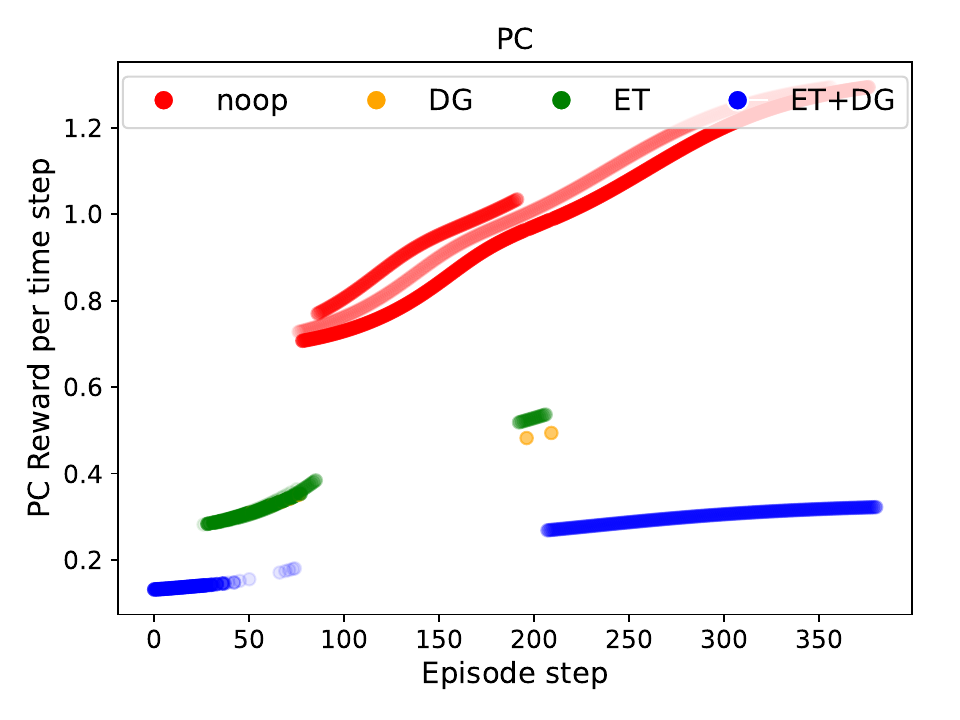}
        \label{fig:pc_ra}
    \end{subfigure}
    \begin{subfigure}[b]{0.32\textwidth}
        \includegraphics[width=\textwidth]{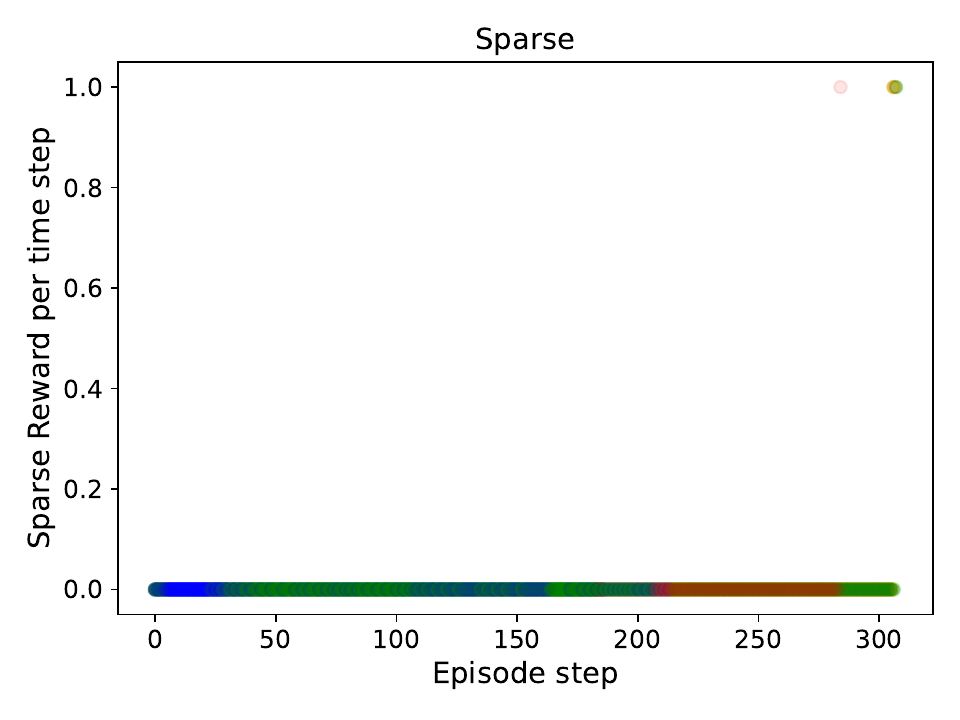}
        \label{fig:sparse_ra}
    \end{subfigure}
    \caption{Reward obtained and action taken per timestep of different sucessful DQN agents trained with different reward functions.}
    \label{fig:rewards_per_action}
\end{figure}

This shows that there are many qualitatively different pathways towards achieving the goal in the AYS environment, and that reward signals can easily be used as a way to embed structure into policy space.
However, finding a strategy for precisely tuning the trajectories such that they may "evolve" in some specific manner is still an open problem, and we believe it to be robust.
Nonetheless, the effect is noticeable across all our experiments.


\section{Further experiments and conclusions}\label{appendix:fexperiments}

 In our experiments we observed significant issues with sensitivity to hyperparameters, which were very difficult to tune. The off-policy agents were much more flexible with learning the environment in different experiments, which is clear from their consistency across the experiments. The on-policy agents were lacking in exploration, which significantly hurt their performance when using the cost reward function.

\begin{wrapfigure}{r}{0.45\textwidth}
  \vspace{-5mm}
  \centering
  \includegraphics[width=0.43\textwidth]{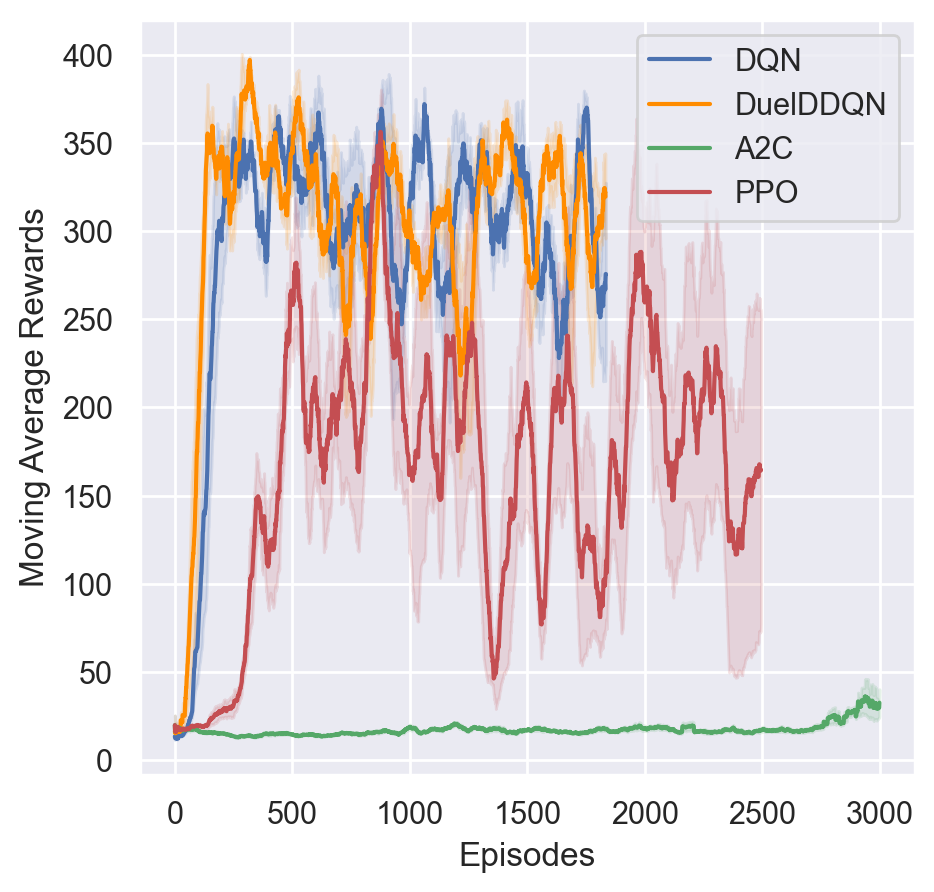}
          \vspace{-4mm}
  \caption{Moving average and standard deviation reward of the agents in the \textit{Noisy} AYS environment. The noise standard deviation is set at $10^{-3}$.}
    \label{fig:noisy_plot}
  \vspace{-7mm}
\end{wrapfigure}

 \paragraph{Comparing AYS to noisy AYS environment}
 We now test injecting Gaussian noise to the parameters of the environment at each new episode. Each new episode is then slightly different. This is significant, as in dynamical systems small changes early on can radically change the outcome, known in mathematics as \textit{chaos}. This tests the agents' robustness to different environment parameters to emulate the fact that such real-world parameters are never perfectly known. The PPO and A2C agents struggle significantly more in this environment but the DQN-based agents are relatively unaffected by the introduction of noise. This is promising as it shows that off-policy agents can learn in this noisy environment. This brings them one step closer to real-world application.

 \paragraph{Comparing fully vs partially observable environments}
 We also test making the environment fully observable to the agent by giving the velocities of the variables as observable features. This enables us to show whether a partially observed environment is a significant hurdle to learning, as the real world is almost certainly only partially observed.
 We find that some of the agents can still learn equivalently well in a partially observed environment if fine-tuned to it (the agents' hyperparameters were optimised for the partially observed environment). We can also contrast how the agent leverages the information from each observable in both environments by analysing the trained neural network parameters. We use SHAP values \citep{Lundberg2017APredictions} as a proxy for feature importance, see \figurename{ \ref{fig:markobddqn}} below. We see that that the state velocity is used more by the agent to infer expected reward.

 \begin{figure}[b]
    \centering
    \includegraphics[scale=0.58]{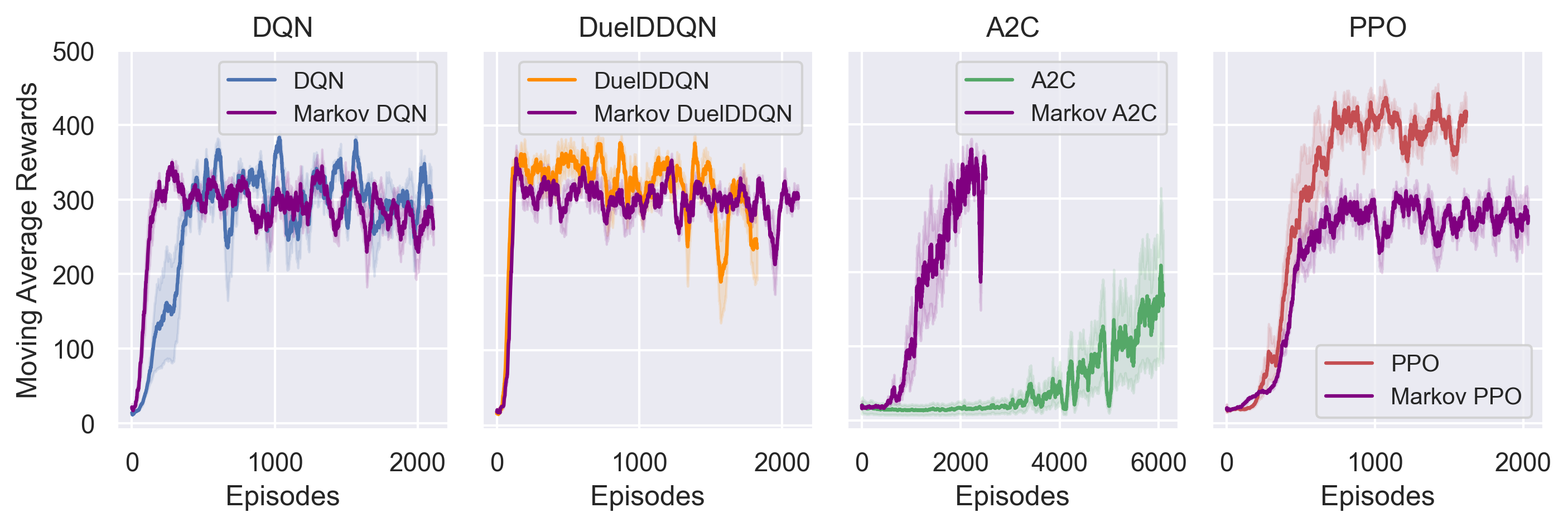}
    \caption{Moving average and standard deviation reward of the four agents with a partially observed environment and with a fully observed one (labelled \textit{Markov}), all trained over 500000 time steps. It is worth noting that the hyperparameters were not re-tuned for the fully observable environment such that we accurately compare identical agents in different environments.}
    \label{fig:markov}
\end{figure}

\begin{figure}[t]
    \centering
    \begin{subfigure}[b]{0.49\textwidth}
    \centering
        \includegraphics[width=\textwidth]{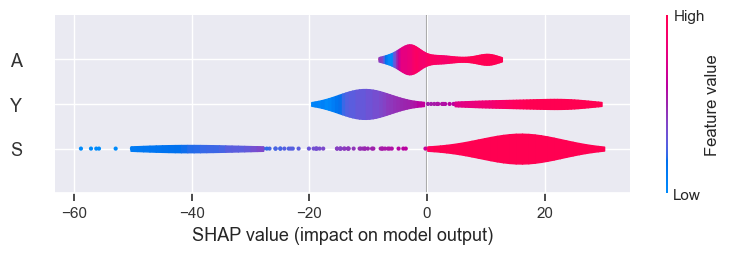}
        \caption{D3QN}
        \label{fig:aysddqn}
    \end{subfigure}
    \begin{subfigure}[b]{0.49\textwidth}
        \includegraphics[width=\textwidth]{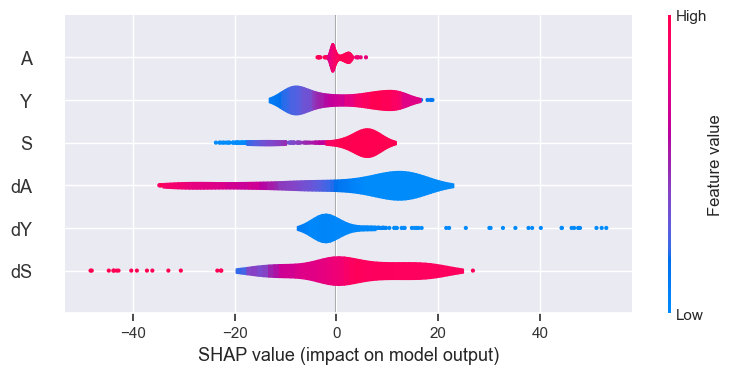}
        \caption{Markov D3QN}
        \label{fig:markobddqn}
    \end{subfigure}
    \caption{SHAP values of two D3QN agents, one trained in the partially observed environment and one trained in the fully observed environment. These plots use 500 states randomly sampled from the replay buffer after training.}
    \label{fig:feat}
\end{figure}

\paragraph{Future extensions} There are many extensions that have not been explored in this work. Changes that were not looked at were changes in the number of actions that can be taken per year, we set this to one throughout this work, but there is no particular reason for this, apart from the easily interpretable idea of one policy per year. In this work, we focused more on the interpretable aspect and thus aimed to leave the fundamental dynamics of the model from \cite{Kittel2017FromManagement} untouched. Additional actions or continuous actions are a clear avenue for probing the environment in different ways. There is also research in \textit{Explainable Artificial Intelligence} (XAI) that could be integrated in this framework, specifically: explainable RL \citep{Juozapaitis2019ExplainableDecomposition,vanderWaa2018ContrastiveConsequences}. This would help with explainability of the agents and interpreting their decisions. Multi-agent RL may also be promising, simulating different drivers of different nations through differentiated reward functions.

\end{document}